\documentclass[final,5p,times,twocolumn]{elsarticle}
\usepackage{graphicx}
\usepackage{hyperref}
\usepackage{graphicx}
\usepackage{wrapfig}
\usepackage{lipsum}
\usepackage{booktabs}
\usepackage{float}
\usepackage{natbib}
\usepackage{caption}
\usepackage{makecell}
\usepackage{color,soul}
\restylefloat{table}
\floatstyle{plaintop}

\begin{document}
\begin{frontmatter}

\title{Advancing Depression Detection on Social Media Platforms Through Fine-Tuned Large Language Models}
\author[1]{Shahid Munir Shah}
\affiliation[1]{organization={Department of Computing, Faculty of Engineering, Science, and Technology, Hamdard University},
            city={Karachi},
            country={Pakistan}}
\ead{shahid.munir@hamdard.edu.pk}
\author[2]{Syeda Anshrah Gillani}
\affiliation[2]{organization={DoaZ},
            country={South Korea}}
\ead{ansharah@hamdard.edu.pk}
\author[3]{Mirza Samad Ahmed Baig}
\affiliation[3]{organization={Danat Fz LLC(owned by Argaam)},
                       country={Pakistan}}
\ead{Mirzasamadcontact@gmail.com}         
\author[4]{Muhammad Aamer Saleem}
\affiliation[4]{organization={Faculty of Engineering, Science, and Technology, Hamdard University},
            city={Karachi},
            country={Pakistan}}
\ead{aamer.saleem@hamdard.edu.pk}
\author[5]{Muhammad Hamzah Siddiqui}
\affiliation[5]{organization={Department of Computing, Faculty of Engineering, Science, and Technology, Hamdard University, Karachi, Pakistan}}
         \ead{Hamzahsiddiqui01@gmail.com}
\begin{abstract}
This study investigates the use of Large Language Models (LLMs) for improved depression detection from users' social media data. Through the use of fine-tuned GPT-3.5 Turbo 1106 and LLaMA2-7B models and a sizable dataset from earlier studies, we were able to identify depressed content in social media posts with a high accuracy of nearly 96.0\%. The comparative analysis of the obtained results with the relevant studies in the literature shows that the proposed fine-tuned LLMs achieved enhanced performance compared to existing state-of-the-art systems. This demonstrates the robustness of LLM-based fine-tuned systems to be used as potential depression detection systems. The study describes the approach in depth, including the parameters used and the fine-tuning procedure, and it addresses the important implications of our results for the early diagnosis of depression on several social media platforms.
\end{abstract}

\begin{highlights}
\item Fine-Tuning of the LLMs (GPT-3.5 Turbo 1106 and LLaMA2-7B)
\item Using the fine-tuned models to detect depression from users' social media data more precisely as compared to the traditional models
\item Achieving improved results compared to state-of-the-art models in the literature
\end{highlights}


\end{frontmatter}


\begin{abstract}
This study investigates how using fine-tuned Large Language Models (LLMs) can improve the identification of depression on social media. Through the use of a fine-tuned GPT-3.5 Turbo 1106 model and LLaMA2-7B and a sizable dataset from earlier studies, we have been able to identify depressed content in social media posts with a nearly 96 percent accuracy. This study describes the approach in depth, including the parameters used and the fine-tuning procedure, and it addresses the important implications of our results for the early diagnosis of depression on several social media platforms.
\end{abstract}

\section{Introduction}
\label{intro}
Depression is a widespread mental health issue that affects millions of individuals worldwide \citep{kim2023automatic}. It significantly lowers the quality of life, productivity, and general well-being. 
It is characterized by persistent melancholy, apathy, and a range of medical and emotional difficulties \cite{DSM5}. Such difficulties to an extreme extent may lead to suicidal ideations and behaviors \cite{ribeiro2018depression}. The World Health Organization (WHO) has particularly emphasized the need for early diagnosis and treatment of depression, highlighting its status as a major global cause of disability \cite{WHO2017}.   
Traditionally, depression is diagnosed using manual self-reports, questionnaires, and testimony from friends and relatives \citep{kessler2000methodological, smith2013diagnosis}. However, these methods are time-consuming and mostly produce unreliable results \citep{hussain2020exploring}. 

Depressed people often remain confined to their homes, avoid social gatherings, and use social media to share their feelings or emotional states through their posts \cite{Naslund2016}. Social media activities by depressed individuals generate a rich data source containing important insights and indicators of depression \cite{Reece2017}. Extracting such insights present valuable opportunities for early identification and treatment of depression before it causes further deterioration \cite{DeChoudhury2013, Birnbaum2020}
Several studies have explored the field of early detection and diagnosis of depression from users' social media data and multiple approaches have been utilized for this purpose. 
It includes conventional Machine Learning (ML) \cite{vasha2023depression, jickson2023machine}, Deep Learning (DL) \cite{tejaswini2024depression, liu2024depression}, Natural Language Processing (NLP) \cite{bokolo2023deep}, and hybrid approaches. Although these techniques have proven effective to some extent, they often lack contextual knowledge and sensitivity, leading to less accurate results \cite{Ernala2019}. Also, issues with the sparsity and the structure of data, the volatile nature of social media discussion, and the intricacy of human emotions have faced challenges in the diagnosis \cite{Chancellor2019}.

Having  
extensive knowledge of natural language and contextual complexity, Large Language Models (LLMs) may provide an acceptable answer to the issues that previous techniques experienced. These models could find linguistic and semantic patterns more strongly associated with depressed moods 
\cite{minaee2024large, hua2024large}
and their capability to go deeper into language, where emotional and psychological processes are more sensitively conveyed make them more suitable candidates to detect depression from users' text data \cite{guo2024large}. 

Given the limitations of the earlier methods and the strong in-context learning ability of LLMs, this research presents the use of fine-tuned LLMs for depression detection through users' social media text data. Fine-tuning of LLMs is essential to fully leverage their capabilities, as they are trained on general-purpose knowledge but require domain-specific expertise \cite{yu2024experimental}. By leveraging their extensive prior knowledge, we aim to fine-tune LLMs for the specific task of recognizing depression. This process improves model performance on certain tasks and addresses ethical problems, offering a more trustworthy tool for mental health monitoring through improving interpretability and decreasing false positives.

Following are the contributions of our research:
\begin{enumerate}
\item Fine-tuning of the LLMs (GPT-3.5 Turbo 1106 \cite{openai2024gpt35turbo} and LLaMA2-7B \cite{yang2023mentallama} to detect depression more precisely from the popular depression dataset \cite{Shen2017IJCAI}.
\item Using the fine-tuned models to detect depression from users' social media data.
    \item Achieving improved results compared to the state of the art.
    \item Achieved 96.0\% accuracy on test data.
\end{enumerate}
The remainder of the paper is organized as follows: 
Section \ref{LR} presents the literature review of approaches employed for automatic depression detection using users' social media data. This includes traditional ML, more recent DL, and NLP-based approaches. Based on the reviewed literature, the gap analysis is also presented in this section. Section \ref{methods} presents the methodology adopted in this research. This includes the detail of the GPT-Turbo 1106 and LLaMA2-7B, their fine-tuning process, dataset description, and model evaluation parameters. Section \ref{results} presents the obtained results and discussion on them. Finally, Section \ref{conclusion} presents the conclusion of the study. 
\section{Literature Review}
\label{LR}
\subsection{Traditional Models}
In literature, different traditional approaches have been proposed utilizing traditional models (ML and DL) to automatically detect depression and other mental illness conditions from users' social media data. Recent research on these approaches is presented below. 

Abdurrahim and Fudholi \cite{fudholi2024mental} proposed a Convolutional Neural Network (CNN) and Bidirectional Long Short-Term Memory (BiLSTM) based model to detect various mental illness conditions from user's Reddit posts. The outcomes of their study demonstrated
that the model effectively identified patterns associated with mental health, resulting in substantial enhancements in
accuracy as compared to the state of the art. 
Oliveira et al. \cite{oliveira2024bag} utilized a transformer-based approach (i.e. Bidirectional Encoder Representations from Transformers (BERT)) for depression and anxiety disorder prediction from users' Twitter posts. Their study proposed that BERT is a superior approach for textual data classification.   
Gorai et al. \cite{gorai2024bert} employed a combination of BERT and an ensemble of multiple CNN for suicide risk prediction from users' Twitter and Reddit data. Their presented approach achieved a considerable performance compared to the existing models in the literature. 
Banna et al. \cite{al2023hybrid} proposed a hybrid approach based on CNN and LSTM models to predict depression from users' Reddit posts. The presented approach showed an excellent performance in predicting depression from users posts. 
Vasha et al. \cite{vasha2023depression} extracted users' Facebook and YouTube posts to compare the performance of several ML algorithms for the task of depression detection. Their proposed models achieved promising results. 
Ansari et al. \cite{Ansari2023} investigated the use of hybrid and ensemble learning methods for automated depression detection tasks through the data collected from different social platforms like Twitter and Reddit. Their study revealed that ensemble models, which combine multiple feature sets, outperformed hybrid models in classifying depressive symptoms from text data.
Alqazaaz et al. \cite{alqazzaz2023deep} compared the performance of traditional ML algorithms with DL-based LSTM networks for the task of mental illness detection through users' twitter data. In their study, LSTM achieved promising results. 

\hfill
\\
Table \ref{LRS} provides the summary of the literature presented above.

\begin{table*}[h]
\centering
  \caption{Summary of traditional methods presented in literature}
  \label{LRS} 
  \begin{tabular}{lll}
  \hline
  \textbf{Reference} &\textbf{Classifier(s) used} & \textbf{Contributions}	 
  \\
  \hline

\citep{fudholi2024mental} & CNN-BiLSTM & \makecell[l]{Modeling with CNN-BiLSTM and Fast Text embedding provided an F1 score and accuracy \\of 85.0\% and 85.0\%, respectively. In comparison to the BiLSTM model, the F1-Score and \\accuracy were both 83.0\%.}\\
\\
\citep{oliveira2024bag}& BERT & \makecell[l]{The proposed model achieved 0.40 F1 score for depression prediction and 0.36 F1 score for \\ Anxiety prediction.}\\
  \\

\citep{gorai2024bert}& BERT, CNN & \makecell[l]{The proposed model performed
better as compared to the recent approaches in detecting \\suicide risk.}\\

\\ 
\citep{al2023hybrid} &CNN, LSTM & \makecell[l]{The employed models achieved overall 99.4\% detection accuracy.}\\
 \\
\citep{vasha2023depression} & \makecell[l]{NB, SVM, RF,\\ DT, LR, KNN} & 
\makecell[l]{Among the used classifiers, SVM achieved the highest accuracy i.e. 75.1\%.}\\

\\
\cite{Ansari2023} & LSTM, LR & \makecell[l]{The study compared the two sets of methods: hybrid and ensemble based on LSTM and LR \\models. The results show that ensemble models outperform the hybrid model by achieving \\75.0\% accuracy and 0.77 F1 scores.}\\
\\
\citep{alqazzaz2023deep} &  \makecell[l]{LR, KNN, SVM,\\ and CNN-LSTM} & \makecell[l]{CNN-LSTM achieved a superior result i.e. 86.2\% detection accuracy as compared to the \\other employed techniques.}\\
\\
\hline
\end{tabular}
\end{table*}

\begin{table*}[h]
\centering
  \caption{Summary of GPT-based methods presented in the literature for mental health applications}
  \label{LRS2} 
  \begin{tabular}{lll}
  \hline
  \textbf{Reference} & \textbf{Model(s) used} & \textbf{Contributions} \\
  \hline
\cite{yang2023mentallama} & \makecell[l]{LLaMA2-7B,\\ LLaMA2-13B} & \makecell[l]{Fine tuned LLaMA2-13B achieved 85.7\% accuracy, while, LLaMA2-7B achieved \\83.9\% accuracy.} \\
\\
\cite{lamichhane2023evaluation} & GPT-3.5 Turbo & \makecell[l]{GPT-3.5 Turbo model achieved F1 scores of 0.73, 0.86, and 0.37 for stress, depression,\\and suicidality detection, respectively.} \\

\\
\cite{wang2024explainable} & \makecell[l]{LLama2-13B-chat, \\SUS-Chat-34B,\\ Neural-chat-7B-v3} & \makecell[l]{Neural-chat-7B-v3 model achieved the best accuracy i.e. 85.8 \% outperforming \\the others employed models.} \\
\\
\cite{danner2023advancing} & GPT-3.5, ChatGPT-4 & \makecell[l]{GPT-3.5 achieved 0.78  F1-score and outperformed the ChatGPT-4 model.} \\
\\
\cite{xu2024mental} & \makecell[l]{Alpaca,\\ Alpaca-LoRA, \\FLAN-
T5, \\GPT-3.5, GPT-4} & \makecell[l]{Fine tuned Alpaca and FLAN-T5 models outperformed the other employed LLaMA2,\\GPT-3.5, and GPT-4 models along with some traditional models in literature by achieving \\balanced accuracies of 72.4\% and 86.8\% respectively.} \\

  \hline
  \end{tabular}
\end{table*}

\subsection{Limitations of the Traditional Approaches}
Although traditional approaches have made great progress in identifying depression from social media, there are still challenges and certain important gaps that this study seeks to fill.

\begin{itemize}
\item \textbf{Contextual Understanding and Nuance}

The contextual and nuanced aspects of the language suggestive of depression are frequently difficult for traditional methods to capture \cite{jamali2023momentary}. Conventional ML makes extensive use of pre-established features, which might not adequately capture the intricacies of depressive expressions \cite{lopez2024machine}. Even though DL models can instantly pick up new features, they are still unable to decipher the emotional nuances and deeper context of messages on social media \cite{figueredo2022early}. 

\item \textbf{Adaptability and Generalization}

A large number of models in use are trained and assessed using datasets that might not accurately reflect the dynamic and varied character of social media conversation. This may result in generalization problems, where models work well on particular datasets but lose accuracy when used on other platforms or with changing linguistic usage \cite{munappy2022data}. 

\item \textbf{Integration Of Domain-Specific Knowledge}

There is a lack of effective integration of mental health knowledge specific to certain topics into computational models to detect depression \cite{islam2024comprehensive}. 

\item \textbf{Scalability and Real-Time Analysis}

The computational resources needed for current methodologies typically limit their scalability and usefulness for real-time analysis of social media platforms \cite{stieglitz2018social}. 
\end{itemize}

\subsection{Capabilities of LLMs to Address Challenges of the Traditional Models}
Compared to traditional approaches, LLMs possess more sophisticated NLP skills and can comprehend text in a more complex way \cite{naveed2023comprehensive}. Furthermore, 
these models are adaptable and their adaptability may be improved by their fine-tuning process \cite{liu2024adamole}. This enables the models to handle a wide range of linguistic expressions and maintain a high degree of accuracy while identifying sentiments such as sadness from various social media texts.
Fine-tuning of the LLMs also improves their capacity to more precisely identify communication indicators associated with mental illnesses such as depression.
Compared to traditional models, LLMs are more scalable solutions. By optimizing their versions, these solutions can be integrated into social media platforms for real-time monitoring and analysis. These can work effectively with Twitter and other social media platforms to provide users with convenience, allowing them to identify early stages of depression based on user posts.


\subsection{Large Language Models}

To overcome the shortcomings faced by the traditional models, several LLM-based models have been proposed in the literature to detect depression from users' text data (refer to Table \ref{LRS2} for the summary of the reviewed literature on the LLM-based models). 

Yang et al. \cite{yang2023mentallama} introduced MentaLLaMA, a large language model fine-tuned for interpretable mental health analysis using social media data. The study focused on enhancing model interpretability through the creation of the Interpretable Mental Health Instruction (IMHI) dataset. The authors demonstrated that their fine-tuned LLaMA-2 models achieved notable results.

Lamichhane \cite{lamichhane2023evaluation} evaluated the performance of LLM-based GPT (GPT-3.5 Turbo) in analyzing three mental illness conditions i.e. stress, depression, and suicidality, using users' social media textual data. The employed model obtained F1 scores of 0.73, 0.86, and 0.37 for stress detection, depression detection, and suicidality detection, respectively. 

Wang et al. \cite{wang2024explainable} presented an explainable approach to detect depression using LLMs (LLama2-13B-chat, SUS-Chat-34B, and Neural-chat-7B-v3) applied to social media data. Their research highlighted the use of LLMs to not only detect depression but also to provide interpretability, making the model's decisions more transparent and understandable in the context of social media data analysis. 

Danner et al. \cite{danner2023advancing} proposed a novel approach by leveraging advanced transformer architecture i.e. BERT and LLMs i.e. GPT-3.5 and ChatGPT-4 for detecting depression. In their research, both transformer-based and LLMs exhibited limited accuracies. The authors suggested that parameter adjustments and data augmentation could enhance accuracies. 

Xu et al. \cite{xu2024mental} introduced Mental-LLM, where they evaluated multiple LLMs including Alpaca, Alpaca-LoRA, FLAN-T5, GPT-3.5, and GPT-4 for predicting various mental health conditions via users online text data. Their study demonstrated that Alpaca and FLAN-T5 models, which underwent instruction fine-tuning, significantly outperformed traditional models such as BERT and even more generalized LLMs like GPT-3.5 in specific tasks. Despite these advancements, the zero-shot and few-shot performance of these models remained limited, with the authors suggesting that further fine-tuning and prompt engineering could enhance their predictive capabilities.

\subsection{Our Proposed Approach}

In this study, we present the use of fine-tuned LLMs i.e. GPT-3.5 Turbo 1106 and LLaMA2-7B for detecting depression through users’
social media text data (refer to section \ref{methods} for detail on fine tuning of the GPT-3.5 Turbo 1106 and LLaMA2-7B models) .

Compared to the studies presented in the literature, we implemented a more refined parameter adjustment approach, achieving significant improvements in accuracy compared to the generalized GPT-3.5 and GPT-4 models employed in previous studies. 
Specifically, our fine-tuning of GPT-3.5 Turbo 1106 resulted in a remarkable accuracy of 96.0 percent, with Precision, Recall, and F1 scores all exceeding 0.95, thereby surpassing the models reported by Xu et al \cite{xu2024mental}. Additionally, our work with LLaMA2-7B yielded an accuracy of 84.0 percent surpassed the model reported by the Yang et al. \cite{yang2023mentallama}. 
The Results achieved in our study demonstrate that GPT-3.5 Turbo 1106 and LLaMA2-7B are better suited for text generation and managing more complex interactions in mental health-related applications. Furthermore, their fine-tuning can lead to more enhanced results.


\section{Methodology}
\label{methods}

\begin{figure*}[hbt]
    \centering
    \begin{minipage}{0.8\textwidth}
        \centering
        \includegraphics[width=\linewidth]{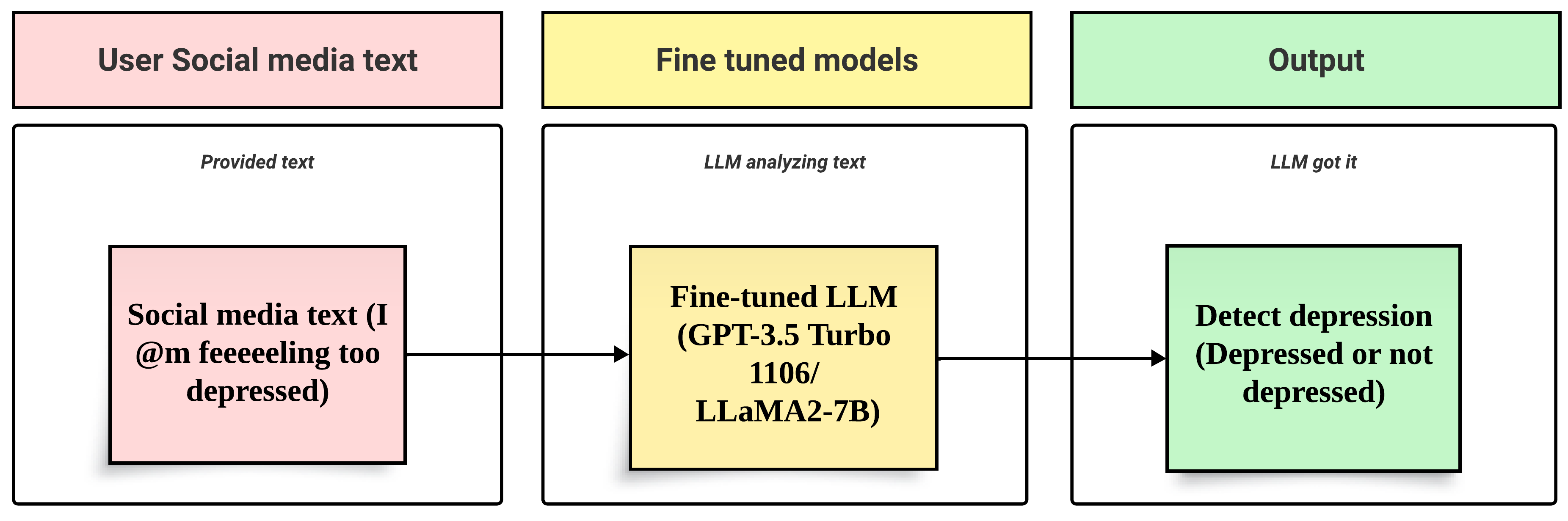}
        \caption{{Methodology adopted}}
        \label{fig1}
    \end{minipage}\hfill 
\end{figure*}

The methodology of this study includes a detailed description of the GPT-3.5 Turbo 1106 model (proprietary model), LLaMA2-7B (open-source), their fine-tuning process, and their use for depression detection from the users' social media data. Figure \ref{fig1} illustrates the adopted methodology. 

According to Figure \ref{fig1}, social media text data is provided to fine-tuned GPT-3.5-Trurbo 1106 \cite{openai2024gpt35turbo} and fine-tuned LLaMA2-7B \cite{touvron2023} models that in turn recognize the provided text as a depressive or non-depressive post. A detailed description of each of these models is provided below:

\subsection{GPT-3.5 Turbo 1106}
GPT-3.5 Turbo 1106 model \cite{openai2024gpt35turbo} is a variant of the GPT series created to facilitate computers in understanding and producing human-like text. It uses efficient transformer architecture effective in processing and creating text. Version 3.5 is the advanced version of GPT-3 where the groundwork of the GPT-4 \cite{openai2023gpt4} is implemented. Table \ref{table1} outlines the specific details of the GPT-3.5 Turbo 1106 and some of its distinctive features have been listed below.


\begin{table*}[h]
   \caption{GPT-3.5 Turbo 1106 model details}
   \label{table1}
    \centering
    \resizebox{\linewidth}{!}{%
    \begin{tabular}{ll}
        \toprule
        \textbf{MODEL} & \textbf{GPT-3.5 Turbo 1106} \\
        \midrule
        \textbf{DESCRIPTION} &  GPT-3.5 Turbo 1106 model is a proprietary advanced language model by OpenAI and is trained on 300 billion tokens. \\
        \textbf{CONTEXT WINDOW} & 16,385 tokens \\
        \textbf{TRAINING DATA} & Up to Sep 2021 \\
        \bottomrule
    \end{tabular}%
    }
\end{table*}



  


\begin{itemize}
    \item \textbf{Pre-training on a Large Corpus} 
    
This model is pre-trained with a large corpus of text data available over the internet. 
This Pre-training on a comprehensive corpus of text data from various sources enables it to understand complex language patterns and context.

\item \textbf{The capacity of the model}
    
This version has a large number of parameters, which are adjustable elements that enable the model to learn from a vast corpus of text data. It is customizable for specific tasks through parameter adjustments, improving accuracy across different applications.

\item \textbf{Fine-Tuning Capabilities}

GPT-3.5 Turbo 1106 has fine-tuning capabilities, and we can fine-tune it on different datasets to perform different tasks. It ranges from general tasks like text completion, and translation to more specialized tasks like depression detection. Fine-tuning of the model adjusts the parameters to suit the pattern and requirement of the task we desire to achieve.

\item \textbf{Advanced Language Understanding}

Because of extensive pre-training capabilities, GPT-3.5 Turbo 1106 is equipped with an appropriate and advanced understanding of the language. It can detect even the tinniest subtle or deep cues and contexts of sad or demotivating posts on social media, which may unnecessarily lead to depression.

\item \textbf{Adaptability}

The model's ability to be fine-tuned makes it highly adaptable to the specific linguistic and contextual nuances of depression-related communication on social media, enhancing its accuracy and effectiveness in this application.

\item \textbf{Scalability} 

GPT-3.5 Turbo 1106 optimized architecture allows deployment in scalable solutions, making it much easier to deploy when integrated with platforms like Facebook, Twitter, etc., for live monitoring of the users of the platform for depression signs and symptoms.

\end{itemize}


\subsection{LLaMA2-7B}
The LLaMA2-7B model excels in combining advanced language processing capabilities with rapid computation speed and adaptable fine-tuning. Its application in depression detection on social media showcases its ability to interpret nuanced language cues effectively and provide real-time insights. With its scalability and robust performance, this model promises significant utility in various critical applications, ensuring its impact across diverse domains. 
Like  GPT-3.5 Turbo 1106, LLaMA2-7B is built on the transformer architecture \cite{46201}, which uses self-attention mechanisms and deep layers to process sequential data effectively. However, it is designed to be more efficient, focusing on optimizing resource usage while maintaining performance across various tasks. Below several distinctive features of LLaMA2-7B have been listed.
\begin{itemize}
    \item \textbf{Pre-training on Diverse Data Sources}

This model is pre-trained on a wide-ranging corpus, which includes not only traditional text data but also specialized datasets. This diverse pre-training enhances its ability to understand complex and varied language patterns, making it highly versatile.

\item \textbf{Parameter Efficiency}

LLaMA2-7B, with its 7 billion parameters, strikes a balance between model size and performance. It is designed to be more parameter-efficient, meaning it can achieve similar or even better results compared to larger models while using fewer resources.

\item \textbf{Increased Throughput}

LLaMA2-7B is optimized for high throughput, making it capable of handling large-scale data processing tasks with reduced computational overhead. This makes it suitable for applications where both speed and resource efficiency are crucial.

\item \textbf{Fine-Tuning Capabilities}

LLaMA2-7B also offers robust fine-tuning capabilities, allowing it to be adapted for specific tasks. Its architecture is designed to quickly learn from task-specific data, improving performance on specialized applications like sentiment analysis, language translation, and more.

\item \textbf{Comprehensive Language Understanding}

Due to its extensive and diverse pre-training, LLaMA2-7B has a nuanced understanding of language, enabling it to detect subtle cues and contextual indicators of depression in social media posts.

\item \textbf{Resource Efficiency}

LLaMA2-7B’s efficient use of parameters and computational resources allows it to perform real-time analysis of social media data without compromising on accuracy, which is essential for live monitoring.

\item \textbf{Adaptability}

With its strong fine-tuning capabilities, LLaMA2-7B can be tailored to recognize the specific linguistic patterns associated with depression on social media, enhancing its effectiveness in this task.

\item \textbf{Scalability}

The model's optimized architecture ensures that it can be deployed at scale, making it suitable for integration into large platforms like Facebook and Twitter for continuous monitoring of depression indicators among users.
Table \ref{lama} provides a detailed overview of the technical specifications of the LLaMA2-7B model.

\end{itemize}

\begin{table*}[h] 
\caption{LLaMA2-7B model details} 
   \label{lama}
    \centering
    \begin{tabular}{ll}
        \toprule
        \textbf{MODEL} & \textbf{LLaMA2-7B} \\
        \midrule
        \textbf{DESCRIPTION} &  LLaMA2-7B is a large language open-source model by META AI and it is trained on 2 trillion tokens \\
        \textbf{CONTEXT WINDOW} & 4096 tokens \\
        \textbf{TRAINING DATA} & Up to Sep 2022 \\
        \bottomrule
    \end{tabular}%
\end{table*}


\subsection{Comparison of GPT-3.5 Turbo 1106 and LLaMA2-7B models}

\begin{figure*}[hbt]
    \centering
    \includegraphics[width=0.7\textwidth]{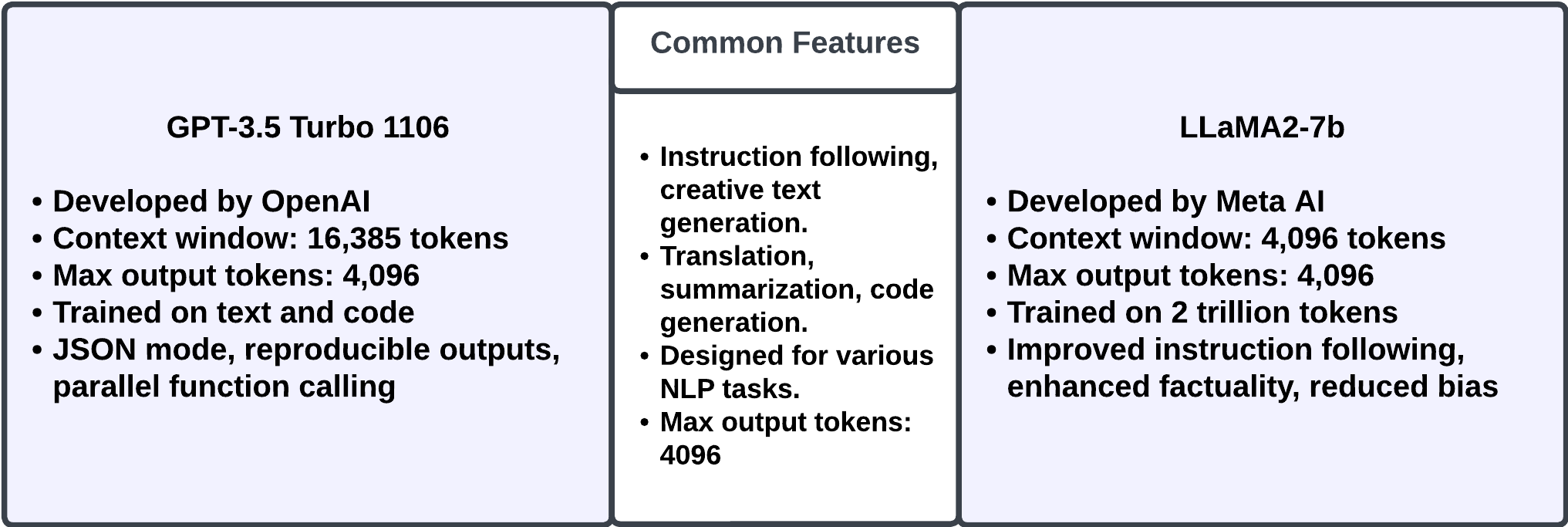} 
    \caption{\textbf{GPT-3.5 Turbo 1106 vs LLaMA2-7B}}
    \label{fig2}
\end{figure*}

Both GPT-3.5 Turbo 1106 and LLaMA2-7B models are designed for a wide range of text-based applications. Along with several common features, these models have different distinctive capabilities to process text applications. Figure \ref{fig2} provides the comparative analysis of the GPT-3.5 Turbo 1106 and  LLaMA2-7B models. As indicated through Figure \ref{fig2}, both the models have been developed by different organizations and have different context window tokens. Furthermore, both the models have trained on different data features. A subtle difference between both models is their reuse capability. GPT-3.5 Turbo 1106 is a closed source model, hence, not freely available to reuse and only a limited number of features can be modified as permitted by the owners. On the other hand, LLaMa2-7B is an open source model, is freely available to reuse and can be modified as per the requirements of users.

\subsection{Experimentation}

\begin{figure*}[htbp]
    \centering
    \includegraphics[width=0.4\textwidth]{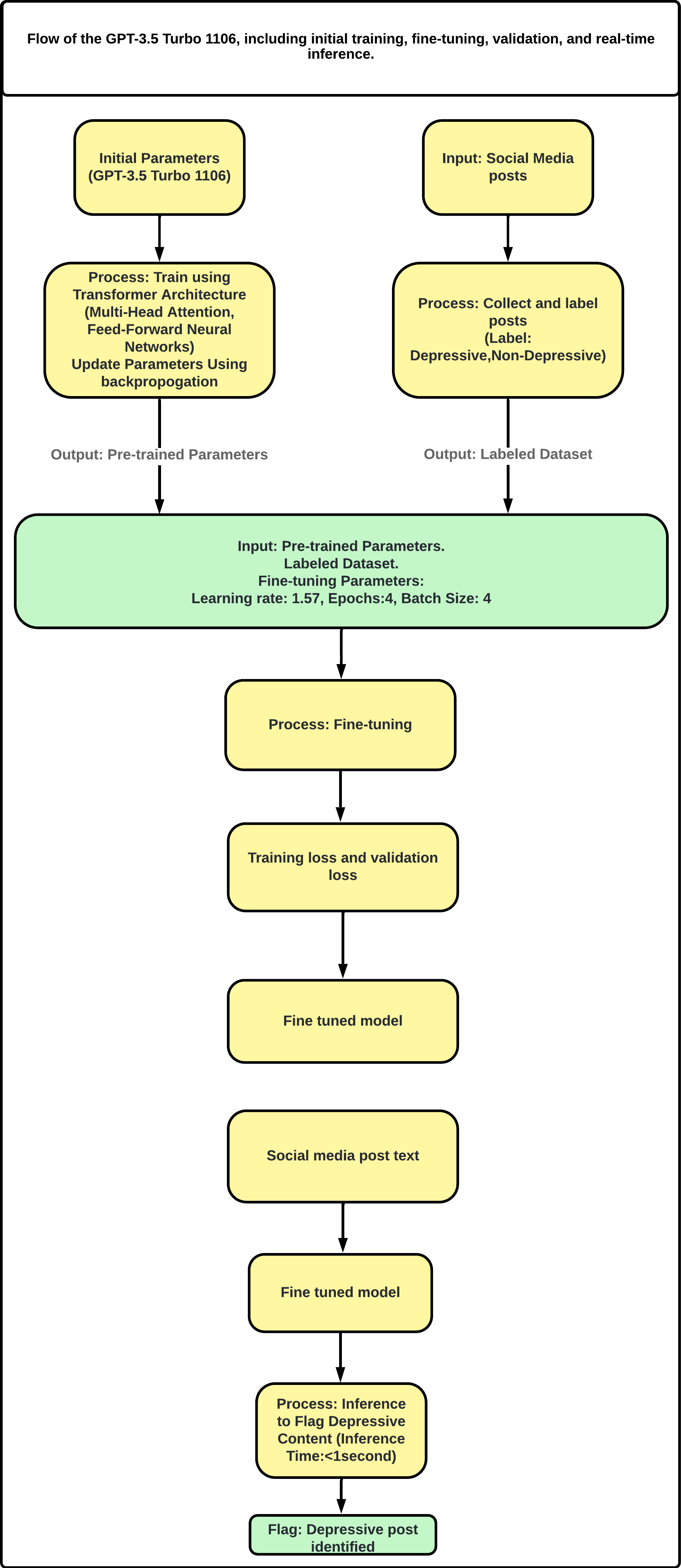} 
    \caption{\textbf{GPT-3.5 Turbo 1106 Training Process and Model Inference.}}
    \label{fig3}
\end{figure*}

\begin{figure*}[htbp]
    \centering
    \includegraphics[width=0.4\textwidth]{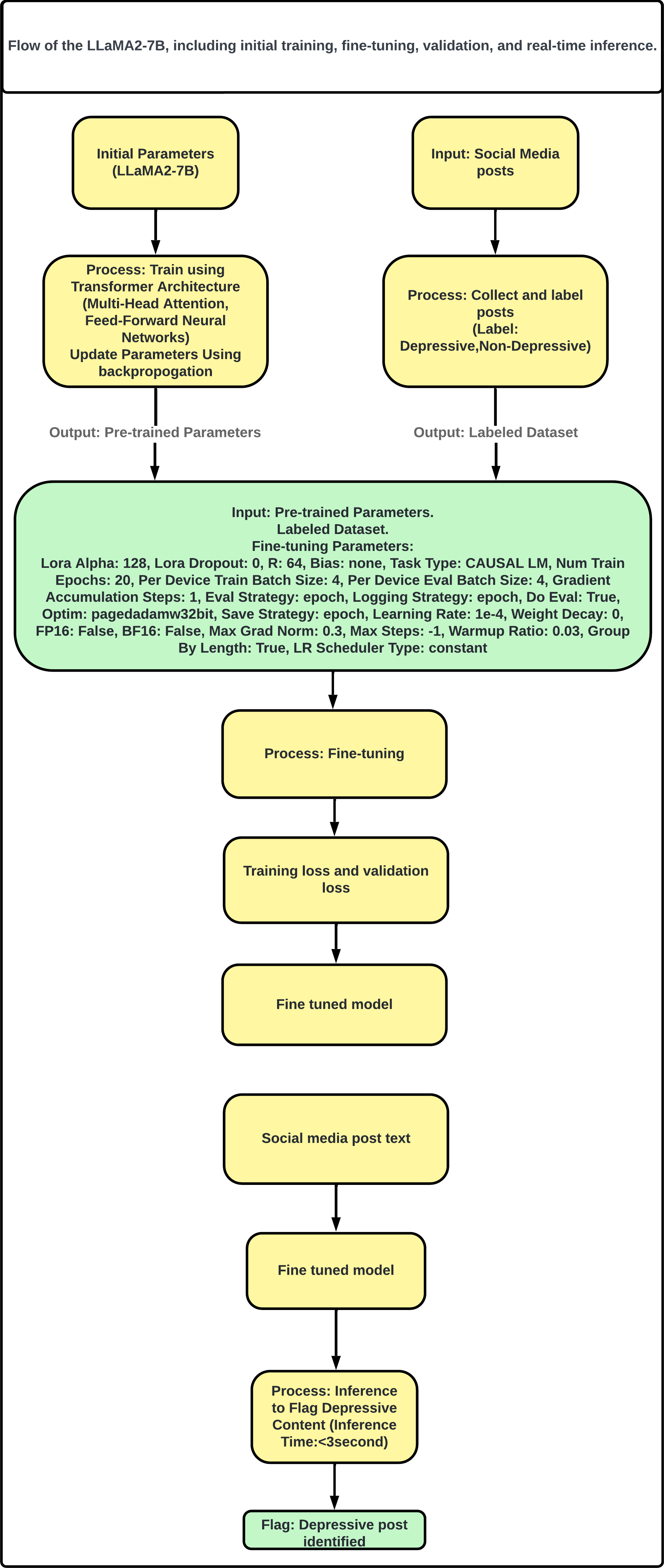} 
    \caption{\textbf{LLaMA2-7B Training Process Model Inference.}}
    \label{fig4}
\end{figure*}

This research utilizes fine-tuned GPT-3.5 Turbo 1106 and LLaMA2-7B for depression detection through users' text data. Figure \ref{fig3} and Figure \ref{fig4} illustrate the complete training processes of GPT-3.5 Turbo 1106 and LLaMA2-7B models respectively. According to these Figures, the training process of both the models is almost similar. During models training, initially, pre-processed and labeled data is provided to the models and then the models are fine-tuned by refining their parameters. Gradient Descent algorithm is used to fine tune GPT-3.5 Turbo 1106, whereas, LoRA configurations are used to fine-tune LLaMA2-7B. After fine tuning and training processes of models, each trained model is tested using test data and evaluated using evaluation parameters like Precision, Recall, F-measure, and Accuracy. 

\subsubsection{Explanation of the Dataset's Origin, Structure, and Relevance to the Study}

\textbf{Dataset Origin}:

To train the employed models, the dataset was used from the research paper named "Depression Detection via Harvesting Social Media: A Multimodal Dictionary Learning Solution" \cite{Shen2017IJCAI}. It was presented at the International Joint Conference on Artificial Intelligence (IJCAI) in 2017. The dataset was built in a way that it could detect depression from users' social media posts. The dataset was collected from Twitter using various APIs of high-grade quality and global ubiquity.
\\
\hfill
The dataset comprises three distinct subsets, labeled as D1, D2, and D3, each serving a unique purpose in the context of depression detection.
\\
\hfill
\textbf{Depression Dataset D1}: 

This subset includes data from Twitter users labeled as depressed based on their anchor tweets between 2009 and 2016 that explicitly mentioned a diagnosis of depression using specific phrases such as "I'm diagnosed with depression" This direct approach ensures a high level of confidence in the depression labels assigned to this dataset.
\\
\hfill
\textbf{Non-Depression Dataset D2}: 

In contrast to D1, D2 consists of users' tweets labeled as non-depressed, specifically, which did not contain any references to "depress" from December 2016. This dataset provides a clear delineation between depressed and non-depressed users based on the absence of depressive language in their tweets.
\\
\hfill
\textbf{Depression-candidate Dataset D3}: 

Recognizing the limitations of D1's size, D3 was constructed to include a larger pool of potential depressed users based on more loosely defined criteria. Users in D3 had anchor tweets from December 2016 containing the term "depress" though this set includes a higher degree of noise due to the less stringent inclusion criteria.
\\
\hfill
\textbf{Features}:

The datasets were supplemented with six feature groups intended to capture both offline behaviors as described by clinical depression criteria and online behaviors seen on social media platforms. These features include Social Network Characteristics, Users Profile Features, Visual Features, Emotional Features, Topic-Level Features, and Domain-Specific Features. This extensive features set attempts to give a multidimensional perspective of each user, which will improve the model's capacity to predict depressed tendencies.
\\
\hfill
\textbf{Sentimental Emoji Library}:

To further refine the dataset, emojis were processed and categorized based on their sentiments, creating a sentimental emoji library. This aspect of the dataset acknowledges the role of emojis in conveying emotions and sentiments in online communication, an important factor in analyzing social media content for signs of depression.
\\
\hfill
\textbf{Relevance to the Study}:

20,000 labeled instances from this extensive dataset were used for training to optimize the GPT-3.5 Turbo 1106 model for depression identification. This ensures a robust learning process that takes into account the complexity and diversity of depressive emotions on social media. An accurate evaluation of the model's performance was made possible by the testing of an extra 20,000 samples that were not included in the training set. The dataset played a crucial part in this research endeavor because of its structure and depth, which, when paired with the model's sophisticated language processing capabilities, allowed for an unparalleled 96.0 Percent accuracy in the detection of depression.

\subsubsection{Comprehensive overview of the fine-tuning process}

Modification of the parameters (weights and biases) of the pre-trained models like  GPT-3.5 Turbo 1106 and LLaMA2-7B to perform a particular task, such as depression detection from social media posts, is known as fine-tuning of the models. To detect depression from users' social media data, the fine-tuning of the employed models is essential. It is to make the models capable of detecting the patterns of the language used in social media posts related to depressive attitudes.
The fine-tuning process of GPT-3.5 Turbo 1106 and LLaMA2-7B is explained below.

\subsubsection{Fine-Tuning of GPT-3.5 Turbo 1106}
    
The following steps have been taken to fine-tune the GPT-3.5 Turbo 1106 model for improving its ability to detect depressive words and phrases from social media posts:


\begin{itemize}

\item \textbf{Parameters Update:}
    
The following parameters of the model have been updated:
\begin{enumerate}
\item \textbf{Epochs}:

The model has been fine-tuned for 4 epochs (epochs refers to one-pass through the training dataset). It means that during training, the model completed 4 passes completely. This number has been chosen to set the sequence length of the dataset 
well suited to the main body of GPT-3 to which the model is pre-trained.

\item \textbf{Batch Size}: 

The batch size was kept at 4. This means that the data will be distributed among four items during training. The smaller data size is selected to encounter frequent changes to the accuracy of the model. However, the data size is still large enough to provide better termination guarantees. The smaller batch size increases the likelihood that the learning process will arrest at a local minimum.

\item \textbf{Learning Rate Multiplier}: 

The multiplier of 1.57 was used to fine-tune the learning rate in the desired direction. This component was utilized to govern the learning rate, which determines how frequently the model’s weights are updated during training. 
\end{enumerate}

\item \textbf{Update Process:}

 The updatee process involves adjusting the model pre-trained weights 
 through the back-propagation propagation algorithm, enabling the model to better perform on the target task.

\item \textbf{Validation Process:}

When the fine-tuning process is finished, the fine-tuned model is sent to a validation set, having a look at the performance of
the model and confirming that any over-fitting has not begun. 
\end{itemize}

\subsubsection{Fine-Tuning of LLaMA2-7B}
The fine-tuning of LLaMA2-7B utilized LoRA \cite{hu2021loralowrankadaptationlarge} configurations to adapt the pre-trained model to better detect depressive content from the dataset. Like the GPT-3.5 Turbo 1106, fine-tuning of LLaMA2-7B consists of the following steps:
\begin{itemize}
    \item \textbf{Parameters Update:}

Table \ref{tab:training_and_configuration_parameters} provides a list of the updated values of the Lora parameters that have been used in our study. 

\begin{table*}[h!]
\centering
\caption{Training and configuration parameters of LoRA}
\begin{tabular}{@{}ll@{}}
\toprule
\textbf{Parameter} & \textbf{Value} \\ \midrule
Lora Alpha & 128 \\
Lora Dropout & 0 \\
R & 64 \\
Bias & none \\
Task Type & CAUSAL LM \\
Num Train Epochs & 20 \\
Per Device Train Batch Size & 4 \\
Per Device Eval Batch Size & 4 \\
Gradient Accumulation Steps & 1 \\
Eval Strategy & epoch \\
Logging Strategy & epoch \\
Do Eval & True \\
Optim & paged\_adamw\_32bit \\
Save Strategy & epoch \\
Learning Rate & 1e-4 \\
Weight Decay & 0 \\
FP16 & False \\
BF16 & False \\
Max Grad Norm & 0.3 \\
Max Steps & -1 \\
Warmup Ratio & 0.03 \\
Group By Length & True \\
LR Scheduler Type & constant \\ \bottomrule
\end{tabular}
\label{tab:training_and_configuration_parameters}
\end{table*}

\item \textbf{Update Process:} 

Fine-tuning with PEFT LoRA \cite{hu2021loralowrankadaptationlarge} involves injecting low-rank adapters into the model layers, allowing for efficient parameter updates without altering the entire model. This method updates only a small subset of the model weights, making it highly efficient and scalable.

\item \textbf{Validation Process:} 

After the fine-tuning phase, the model undergoes rigorous testing against a validation set to ensure its ability to accurately identify depressive content without model over-fitting.
\end{itemize}

\subsection{Model Evaluation}

The performance of the employed models has been evaluated using several important evaluation parameters that include Precision, Recall, F-Score, and Accuracy values. Each of these parameter is briefly explained below: 
\begin{itemize}
    \item \textbf{Precision}:

Precision refers to the proportion of the predicted positives of the model that are positive. Mathematically,
\begin{equation}
 Precision = \frac{TP}{TP+FP}   
\end{equation}

\item \textbf{Recall}:

Recall refers to the proportion of the actual positives correctly classified as positives by the model. Mathematically,
\begin{equation}
 Recall = \frac{TP}{TP+FN}   
\end{equation}

\item \textbf{F-Score}:

F-Score is the harmonic mean of the Precision and Recall. Mathematically, 
\begin{equation}
F-Score = \frac{2*TP}{2*TP+FP+FN} 
\end{equation}
\item \textbf{Accuracy}:

Accuracy refers to the proportion of all the correct classifications of the model, Mathematically,
\begin{equation}
Accuracy = \frac{TP+TN}{TP+TN+FP+FN}
\end{equation}

\end{itemize}
In the above all equations:

TP, TN, FP, and FN are the values taken from the confusion matrix. Where; 

\textbf{TP (True Positives)}: represents the actual positives predicted as positive by the model, 

\textbf{TN (True Negatives)}: represents the actual negatives predicted as negative by the model, 

\textbf{FP (False Positives)}: represents the predicted positives by the model, which are negatives, and 

\textbf{FN (False Negatives)}: represents the predicted negatives by the model, which are positives. 

\section{Results and Discussion}
\label{results}
\subsection{Obtained Results and Discussion on them}

The results achieved in this study have been presented in Table \ref{tab:model_performance}, which indicate that the employed models i.e. fine-tuned GPT-3.5 Turbo 1106 and LLaMA2-7B achieved 96.0\% and 84.0\% accuracies respectively in detecting depression from the users' text data. 
Furthermore, the models also achieved reasonably high values (more than 0.8 and 0.9) of Precision, Recall, and F1-scores, indicating their good performance. 

\begin{figure*}[htbp]
    \centering
    \includegraphics[width=0.7\textwidth]{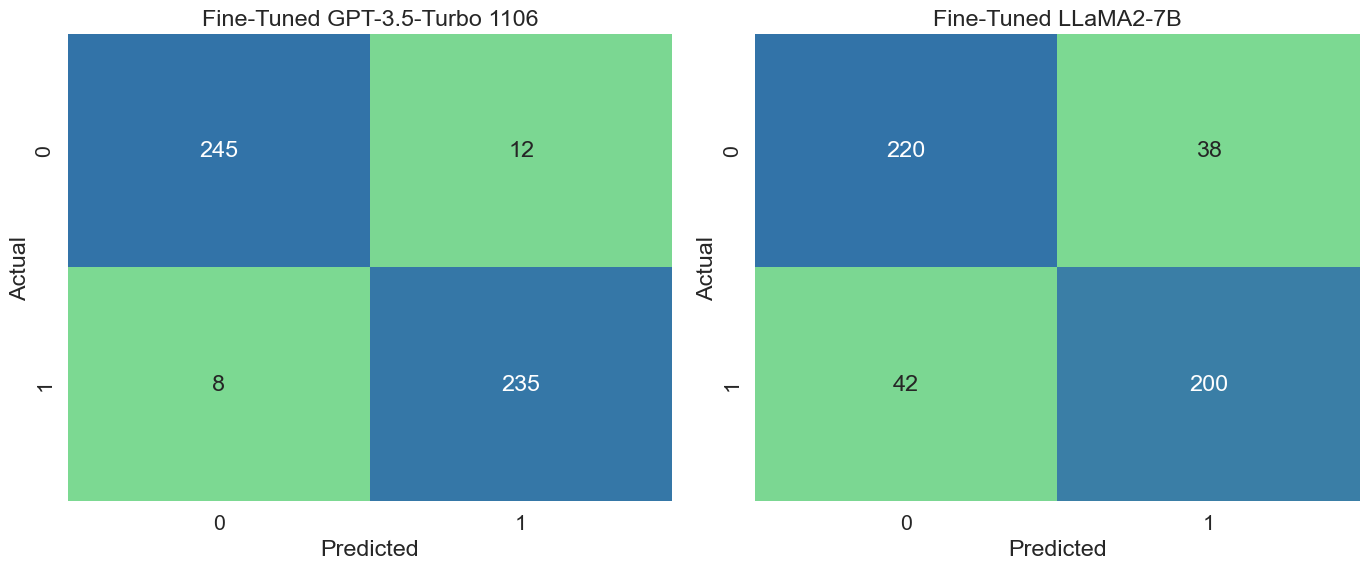} 
    \caption{\textbf{Confusion matrices of fine-tuned GPT-3.5 Turbo 1106 and Fine-Tuned LLaMA2-7B}}
    \label{fig5}
\end{figure*}

\begin{table*}[h!]
    \centering
      \caption{Comparative analysis of the results}
    \label{tab:model_performance}
    \begin{tabular}{llllll}
        \toprule
        Model & accuracy (\%)  & Precision & Recall & F1-Score \\
        \midrule
        Fine-tuned  GPT-3.5 Turbo 1106  & 96            & 0.954    & 0.968  & 0.960 \\
        Fine-tuned LLaMA2-7B        & 84             & 0.852    & 0.840  & 0.846 \\
        GPT-3.5 Turbo\cite{openai2024gpt35turbo}             & 68             & 0.640    & 0.696  & 0.667 \\
        GPT-4 \cite{openai2023gpt4}                     & 74            & 0.745    & 0.745  & 0.745 \\
        GPT-4 Turbo\cite{openai2024gpt4turbo}               & 74           & 0.755    & 0.725  & 0.741 \\
        GPT-4 Omni\cite{openai2024hellogpt4o}                & 72            & 0.700    & 0.729  & 0.715 \\
        Gemini \cite{gemini2023}                   & 62            & 0.604    & 0.604  & 0.604 \\
        \bottomrule
    \end{tabular}
\end{table*}
To further highlight the achieved values of the evaluation parameters, the Confusion Matrices of the 
models have also been obtained and presented in Figure \ref{fig5}. The Confusion Matrices of the employed models indicate that both the models achieved high values in true class prediction, which are the indicators of the better-achieved accuracy values.





\begin{figure*}[htbp]
    \centering
    \includegraphics[width=0.6\textwidth]{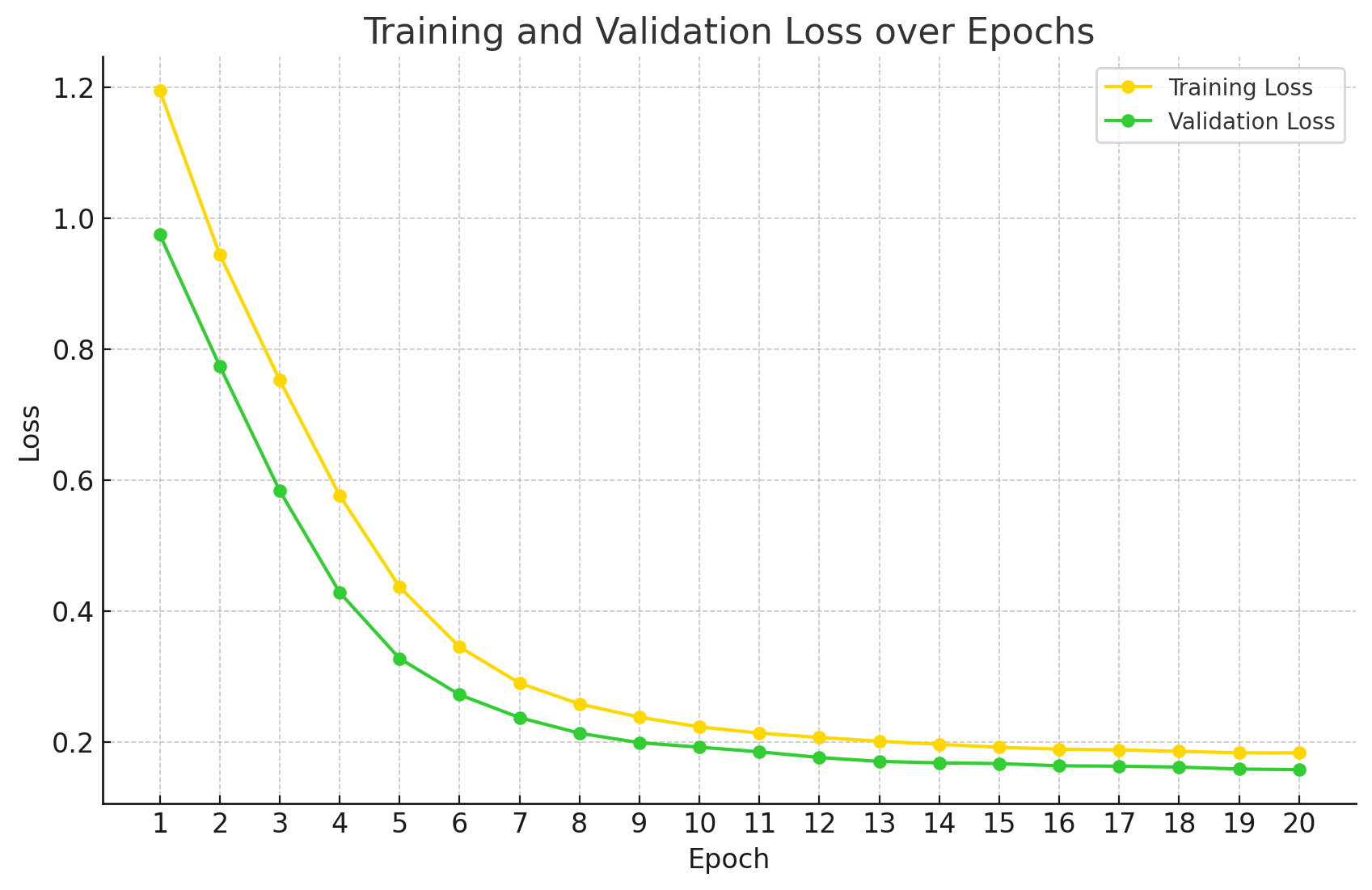}
    \caption{\textbf{Training and validation loss of fine-tuned LLaMA2-7B}}
    \label{loss-Lamma}
\end{figure*}

\begin{figure*}[htbp]
    \centering
    \includegraphics[width=0.6\textwidth]{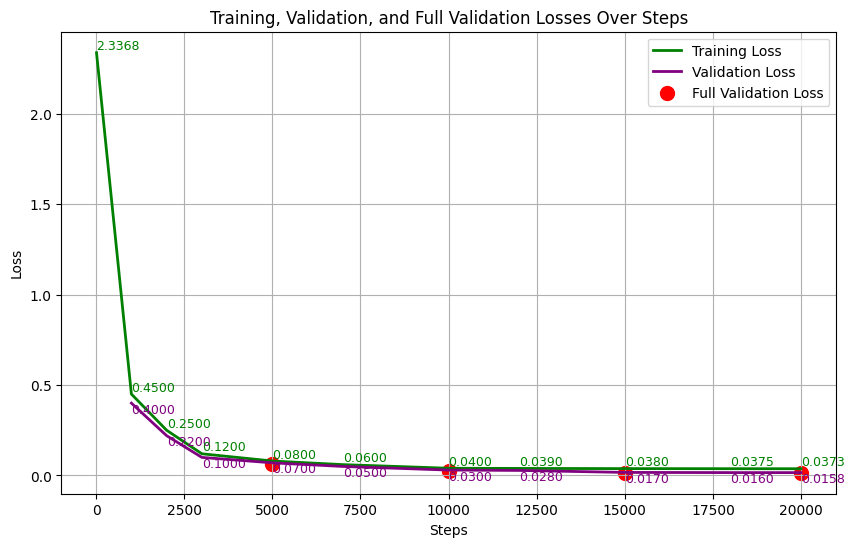}
    \caption{\textbf{Training and validation loss of fine-tuned GPT-3.5 Turbo 1106}}
    \label{loss-gpt}
\end{figure*}

Figures \ref{loss-Lamma} and \ref{loss-gpt} present the training and validation losses of the fine-tuned LLaMA2-7B and GPT-3.5 Turbo 1106 respectively.  

According to Figure \ref{loss-Lamma}, we obtained a training loss of 0.18 and a validation loss of 0.16 after 20 epochs. In this figure, the yellow line represents the training loss, the green line represents the validation loss, the x-axis denotes the number of steps, and the y-axis indicates the loss values. Full validation is obtained after each successful epoch cycle.

According to Figure \ref{loss-gpt}, we obtained a training loss of 0.034, validation loss of 0.016, and, full validation loss of 0.153.

\begin{figure*}[htbp]
    \centering
    \includegraphics[width=0.6\textwidth]{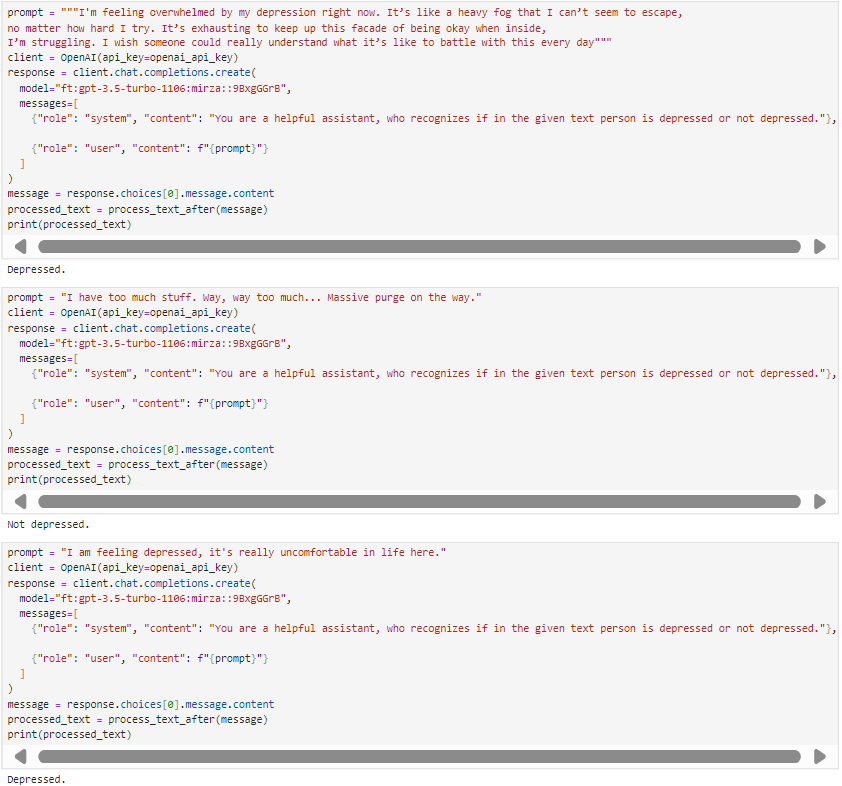}
    \caption{\textbf{Inference produced by the fine-tuned GPT-3.5 Turbo 1106}}
    \label{fig:usage_demo}
\end{figure*}

\begin{figure*}[htbp]
    \centering
    \includegraphics[width=0.9\textwidth]{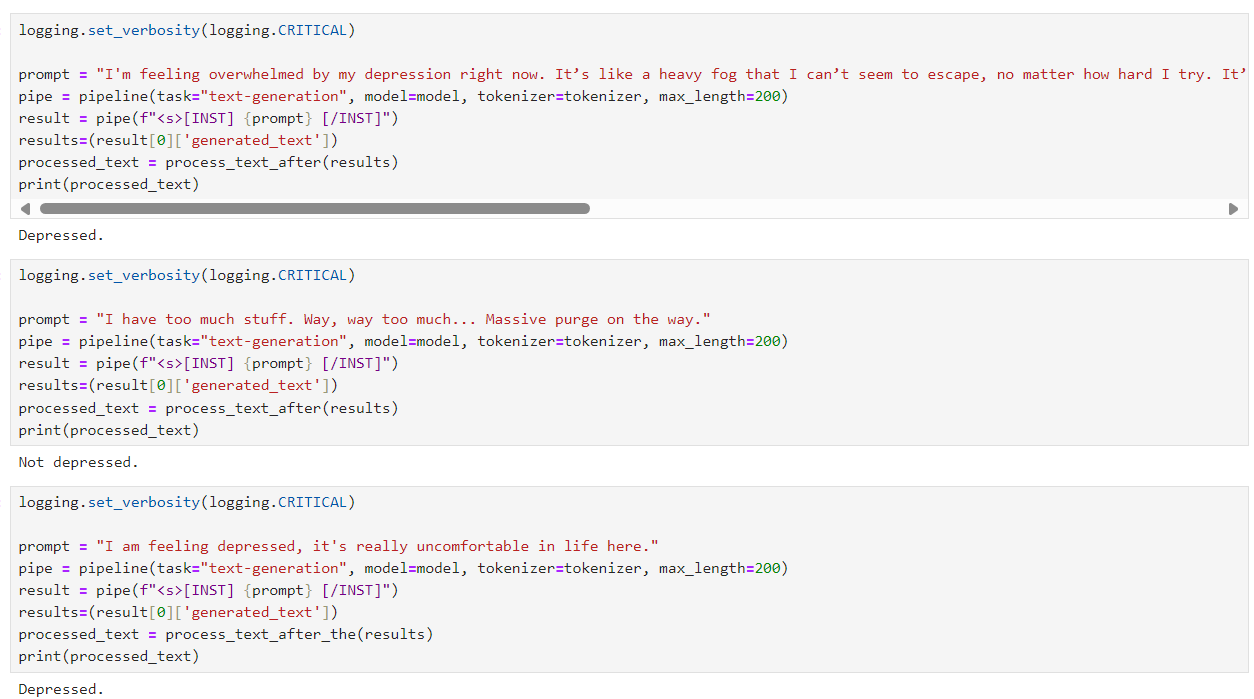}
    \caption{\textbf{Inference produced by the fine-tuned LLaMA2-7B model}}
    \label{fig:usage_dema2}
\end{figure*}

Figures \ref{fig:usage_demo} and \ref{fig:usage_dema2} provide the inference generation of fine-tuned models, which indicates that the models doing well in accurately inferring the depressive textual posts from the users.  

\subsection{Comparative Analysis of Results}

Table \ref{tab:model_performance} provides a comparative analysis of the achieved results of the proposed models and the other existing relevant LLMs in the literature. 
A detail of the comparative analysis of the results in terms of accuracy, precision, Recall, and F1-score values is provided below.
\begin{itemize}
    \item \textbf{Accuracy Comparison:}
    
Fine-Tuned Gpt-3.5 Turbo achieved 96.0\% accuracy and fine-tuned LLaMA2-7B achieved 84.0\% accuracy, indicating great prediction capabilities of the employed models on the given dataset.
In contrast, previous Models generally exhibited high accuracy but fell short of 96.0\%, indicating occasional miss-classifications.

\item \textbf{Precision and Recall Comparison:}

Both Precision and Recall matrices of the employed models have reached their theoretical maxima, demonstrating the models' ability not only to correctly identify positive instances but also to minimize false positives.
In contrast, while achieving high Precision and Recall, previous models exhibited a trade-off between the two, struggling to optimize one without sacrificing the other.

\item \textbf{F1-Score Comparison:}

With perfect Precision and Recall, the employed models also achieved the maximum value of the F1 score i.e. the harmonic mean of the two. This underscores the models' balanced performance in both aspects.
In contrast, the previous models, due to the Precision-Recall trade-off, often had lower F1 scores, indicating less balanced performance.



\end{itemize}

\subsection{Model’s limitations, potential biases}

While the proposed model has demonstrated exceptional performance, it's crucial to consider its limitations, potential biases, and areas for future research, especially in the context of implementing such a model on social media platforms like Twitter for early depression detection.

\subsubsection{Limitations}

\begin{itemize}
    \item \textbf{Data Diversity and Volume:}

     The models' training could have been limited to a specific dataset, which could limit their generalizability across diverse demographics and cultural contexts. There are various languages, slangs, and forms of expression used on social media, and the training data may not have been enough to cover all of them.
   
    \item \textbf{Contextual Understanding:}
    
     The context derived is often based on current events, memes, or cultural references on the internet. The model might not be able to understand the context fully after some months or years and infer wrong meanings. 

    \item \textbf{Dynamic Nature of Language:}
    
    The usage of languages in social media is always dynamic. There are new forms of slang, abbreviations, and symbols coming up almost constantly, most of which the model might not know.
    
\end{itemize}

\subsubsection{Potential Biases}

\begin{itemize}
    \item \textbf{Sampling Bias:}
    
    The training dataset, which is used to fit the model, might not be worldwide sampled. For instance, if the post of depression is studied only from a specific region, age group, or people and is not used for the representation of sampling purposes then generally, the model doesn't work well on the post from outside of this group.

    \item \textbf{Confirmation Bias:} 
    
    While identifying depression, there might be a crucial point of confirmation bias. It means that the models may focus on the signals that support depression rather than giving away other points that contrast such cases. Also, in complex cases, the model tends to pick signals that support depression.

\end{itemize}

\section{Conclusion and Future Directions}
\label{conclusion}
\subsection{conclusion}

This study presented the use of fine-tuned LLMs i.e. GPT-3.5 Turbo 1106 and LLaMA2-7B for detecting depression
through users’ social media text data. Using LLMs, particularly the GPT-3.5 Turbo 1106 and LLaMA2-7B models, is a pioneering approach for recognizing depression from users' social media data. The employed models were trained on a popular depression-related dataset \cite{Shen2017IJCAI} to infer depression from users' social media posts.
\\
\hfill
The following are the key findings of this research:

\begin{itemize}
     \item The employed fine-tuned models i.e. GPT-3.5 Turbo 1106 and LLaMA2-7B achieved excellent results i.e. 96.0 \% and 84.0\% accuracies respectively, and efficiently detected depression from users' posts.
    \item The models also achieved excellent values of the other evaluation parameters i.e. Precision, Recall, and F-Score values, indicating the best performance of the models.
    \item Comparative analysis of the achieved results demonstrated that the employed fine-tuned models outperformed various related models in the literature. 
\end{itemize}

The Results achieved in our study indicate that GPT-3.5 Turbo 1106 and LLaMA2-
7B are better suited for text generation and managing more complex interactions in mental health-related applications. Furthermore, their fine-tuning can lead to more enhanced results.

The results of this research extends beyond the domain of the study. For example, it provides ways to practically integrate the monitoring of mental health on the social media infrastructures to revolutionize how these platforms can give support and troubleshooting in ways like early intervention and connecting with mental resources as well.

In summary, this study achieved an accurate detection rate of depression on social media platforms but also highlighted how the LLMs can transform the public health and wellness domain in a big way. Given the findings and accuracy of these advanced AI solutions, there lies an urgent need for these to be integrated into mental health initiatives.

\subsection{Future Directions}

The future work can focus on improving the model's understanding of the context of a social media post, incorporate world knowledge, and generalize the model to adapt to the dynamic nature of online communication by fine-tuning our fine-tuned model after some months or year on current slangs and trends, etc.

Another future work can focus on making the model adaptable to the new trends in the language on social media in real-time to increase its applicability and use on social media platforms in the long run.

It is crucial to ensure that real-world implementations and the developed models do not raise ethical concerns, particularly if social media platforms themselves deploy them in real time by asking user permission. If this is not possible, we must obtain user consent and ensure users' privacy is protected. In future work, we should explore the ethical implications based on the progress we have made so far and develop a framework to address these issues.

The detection of potential biases and the mitigation algorithm is instrumental in ensuring that the model's predictions are as equitable and unbiased as possible. In future work, we can develop more advanced bias detection and mitigation techniques to further guard against unfair treatment for all groups in the distribution.

To improve prediction accuracy and fairness, human judgment could be incorporated into the model for complex or ambiguous cases. In future systems, it would be useful to combine AI-based decisions with human assessments.

 \bibliographystyle{unsrt} 
 \bibliography{example}

\end{document}